\def\year{2020}\relax
\documentclass[letterpaper]{article} 
\usepackage{aaai20}  
\usepackage{times}  
\usepackage{helvet} 
\usepackage{courier}  
\usepackage[hyphens]{url}  
\usepackage{graphicx} 
\usepackage{times}
\usepackage{helvet}
\usepackage{courier}
\usepackage{url}  
\usepackage{amstext,amsmath,amssymb}
\usepackage{algorithm}
\usepackage{algorithmic}
\usepackage{graphicx}
\usepackage{xcolor}
\usepackage{hyperref}       
\hypersetup{
pagebackref=true,%
colorlinks=true,%
pdftex
}
\usepackage{booktabs}
\usepackage{multirow}
\usepackage[font=footnotesize]{subcaption}
\usepackage{microtype}
\usepackage{stfloats}
\usepackage{graphicx}  
\title{Generative Exploration and Exploitation}
\author{Jiechuan Jiang \\
            Peking University \\
	        \small jiechuan.jiang@pku.edu.cn
\And 
	Zongqing Lu\thanks{Corresponding author} \\
	Peking University \\
	\small zongqing.lu@pku.edu.cn 
}

\begin{document}
\maketitle
 	
\begin{abstract}
Sparse reward is one of the biggest challenges in reinforcement learning (RL). In this paper, we propose a novel method called \textit{Generative Exploration and Exploitation} (GENE) to overcome sparse reward. GENE automatically generates start states to encourage the agent to explore the environment and to exploit received reward signals. GENE can adaptively tradeoff between exploration and exploitation according to the varying distributions of states experienced by the agent as the learning progresses. GENE relies on no prior knowledge about the environment and can be combined with any RL algorithm, no matter on-policy or off-policy, single-agent or multi-agent. Empirically, we demonstrate that GENE significantly outperforms existing methods in three tasks with only binary rewards, including Maze, Maze Ant, and Cooperative Navigation. Ablation studies verify the emergence of progressive exploration and automatic reversing.
\end{abstract}
 	
\section{Introduction}
 	
Deep reinforcement learning (RL) has achieved great success in many sequential decision-making problems, such as Atari games \cite{mnih2015human}, Go \cite{silver2016mastering,silver2017mastering}, and robotic tasks \cite{levine2016end,duan2016benchmarking}. However, a common challenge in many real-world applications is the reward is extremely sparse or only binary. For example, in goal-based tasks, the agent can only receive the reward when it reaches the goal. Nevertheless, the goal is usually hard to reach via random exploration, such as $\epsilon$-greedy and Gaussian noise. Domain-specific knowledge can be used to construct a shaped reward function to guide the policy optimization. However, it often biases the policy in a suboptimal direction, and more importantly domain-specific knowledge is unavailable in many cases.

Some exploration methods have been proposed to address sparse reward. A method family quantifies the novelty of the state and takes it as the intrinsic reward to encourage the agent to explore new states, \textit{e.g.}, count-based exploration \cite{bellemare2016unifying,ostrovski2017count} and curiosity-driven exploration \cite{pathak2017curiosity,burda2018large,burda2018exploration}.
However, intrinsic reward leads to deviation from the true target and causes the learning process detoured and unstable. 
Some methods set additional goals for exploration.
Goal GAN \cite{florensa2018automatic} generates different goals at the appropriate level of difficulty for the agent. HER \cite{andrychowicz2017hindsight} replays each episode with a different goal sampled from the buffer rather than the original one to be achieved. 
However, driven by random exploration the agent still rarely obtains a real reward signal. 
 	
Changing start state distribution has been considered to accelerate learning.
Appropriate start states can improve the policy training and performance, which has been proven theoretically by \cite{kearns2002sparse}. 
Some works adopt the concept of reversing \cite{florensa2017reverse,goyal2018recall}, gradually learning to reach the goal from a set of start states increasingly far from the goal. 
Other researches change the start states by sampling from the states visited by expert demonstrations \cite{nair2018overcoming,resnick2018backplay}. 
However, all these methods require a large amount of prior knowledge and handcrafted designs.

In this paper, we propose a novel method called \textit{Generative Exploration and Exploitation} (GENE) to overcome sparse reward. 
GENE dynamically changes the start states of agent to the generated novel states to encourage the agent to explore the environment or to the generated unskilled states to propel the agent to exploit received reward signals. 
We adopt Variational Autoencoder (VAE) \cite{kingma2013auto} to generate desired states and let the agent play from these states rather than the initial state. 
As the encoder of VAE compresses high-dimensional states into a low-dimensional encoding space, it is easy to estimate the probability density functions (PDFs) of successful states and failed states experienced by the agent via Kernel Density Estimation (KDE) \cite{rosenblatt1956remarks}.
We sample from the distribution to feed into the decoder to reconstruct states. 
By deliberately giving high probability to the state encodings with little difference between these two densities, GENE is able to adaptively guide the agent to explore novel states and to practice at unskilled states as the learning progresses.

GENE can be combined with any RL algorithm, no matter on-policy or off-policy, single-agent or multi-agent. 
Driven by unsupervised VAE and statistical KDE, GENE relies on no prior knowledge and handcrafted designs. 
Like other methods that change start states, GENE requires the start state can be set arbitrarily, which however is feasible in many simulators, \emph{e.g.}, MuJoCo \cite{todorov2012mujoco}, Robotics \cite{brockman2016openai}, MPE \cite{lowe2017multi}, and MAgent \cite{zheng2018magent}. 
Taking advantage of embedding states into a encoding space, GENE is practical and efficient in high-dimensional environments. Moreover, in multi-agent environments with sparse rewards where the search space exponentially increases with the number of agents, GENE can greatly help agents to co-explore the environment.

Empirically, we evaluate GENE in three tasks with binary rewards, including Maze, Maze Ant, and Cooperative Navigation. 
We show that GENE significantly outperforms existing methods in all the three tasks. 
Ablation studies verify the emergence of progressive exploration and automatic reversing, and demonstrate GENE can adaptively tradeoff between exploration and exploitation according to the varying PDFs of successful states and failed states, which is the key to solve these tasks effectively and efficiently.

\section{Related Work}
 	
\subsubsection{Exploration}  

Some methods impel the agent to discover novel states by intrinsic motivation which explains the need to explore the environment. 
These methods fall into two categories: count-based methods and curiosity-driven methods. 
Count-based methods \cite{bellemare2016unifying,ostrovski2017count} directly use or estimate visit counts as an intrinsic reward to guide the agent towards reducing uncertainty.
Curiosity-driven methods \cite{pathak2017curiosity,burda2018large,burda2018exploration} use the prediction error in the learned feature space as the intrinsic reward. 
When facing unfamiliar states, the prediction error becomes high and the agent will receive high intrinsic reward. 
However, the shaped reward is biased and the scale of the intrinsic reward might vary dramatically at different timesteps, which leads to deviation from the true target and causes the learning process detoured and unstable. 
 	
Setting additional goals is another idea for exploration. 
Curriculum learning \cite{bengio2009curriculum,narvekar2019learning} designs a sequence of sub-tasks for the agent to train on, to improve the learning speed or performance on a target task. 
Goal GAN \cite{florensa2018automatic} generates different goals at the appropriate level of difficulty for the agent by adding the label of difficulty level into the GAN's loss function. 
However, it is designed for the multiple-goal situation. 
If there is only one goal in the environment, Goal GAN cannot focus on it, causing the slow learning. 
HER \cite{andrychowicz2017hindsight} is inspired by that one can learn almost as much from achieving an undesired outcome as from the desired one. 
It arbitrarily selects a set of additional goals to replace the original goal. 
However, learning additional goals slows down the learning process, and by random exploration the agent rarely obtains a real reward signal.

\subsubsection{Start State Distribution}

Reversing is the main theme of changing start state distribution. 
Learning from easy states which are close to the goal, to the harder states, until the initial state is solved. 
Reverse Curriculum Generation (RCG) \cite{florensa2017reverse} makes the agent gradually learn to reach the goal from a set of start states which are between the bounds on the success probability. 
However, it requires providing at least one state from which the agent accomplished the task (\emph{i.e.}, reached the goal).
Moreover, RCG is mainly designed for the case where the target state is uniformly distributed over all feasible states. \citeauthor{goyal2018recall} (\citeyear{goyal2018recall}) trained a backtracking model to predict the preceding states that terminate at the given high-reward state. Then the generated traces are used to improve the policy via imitation learning.
\citeauthor{nair2018overcoming} (\citeyear{nair2018overcoming}) reset some training episodes using states from demonstration episodes, and Backplay \cite{resnick2018backplay} samples start states from a window on a demonstration trajectory and slides the window manually. These two methods assume access to expert demonstrations, which are usually unavailable. All the existing methods of changing start states distribution require a large amount of prior knowledge and handcraft designs.

\section{Background}
 	
\subsubsection{Reinforcement Learning}
Consider a scenario where an agent lives in an environment. 
At every timestep \(t\), the agent gets current state \(s_t\) of the environment, takes an action \(a_t\) to interact with the environment, receives a reward \(r_t\), and the environment transitions to the next state. 
Deep RL tries to help the agent learn a policy which maximizes the expected return \(R=\sum_{t=0}^{T}\gamma ^t r_t\). 
The policy can be deterministic \(a_t=\mu(s_t)\) or stochastic \(a_t\sim \pi(\cdot |s_t)\). 
 	
There are two main approaches in RL: policy gradient and Q-learning. 
Policy gradient methods directly adjust the parameters \(\theta\) by maximizing the approximation of \(J(\pi_{\theta})\), \emph{e.g.}, \(J\left ( \theta  \right )=\mathbb{E}_{s\sim p^{\pi},a\sim \pi_{\theta }}\left [ R \right ]\). 
They are almost always on-policy. 
TRPO \cite{schulman2015trust} and PPO \cite{schulman2017proximal} are typical policy gradient methods. 
They all maximize a surrogate objective function which estimates how much \(J(\pi_{\theta})\) will change as a result of the update.  
Q-learning (\emph{e.g.}, DQN) learns a value function \(Q(s,a)\) based on Bellman equation and the action is selected by \(a = \arg \max_a Q(s,a)\). 
Q-learning methods are usually off-policy. 
DDPG \cite{lillicrap2015continuous} learns a Q-function and a deterministic policy, where the Q-function provides the gradient to update the policy. 
MADDPG \cite{lowe2017multi} is an extension of DDPG for multi-agent environments,  making it feasible to train multiple agents acting in a globally coordinated way.   
 	
\subsubsection{Variational Autoencoder} 
VAE consists of an encoder and a decoder. 
The encoder takes a high-dimensional datapoint \(x\) as the input and outputs parameters to \(q_{\theta}(z|x)\). 
A constraint on the encoder forces the encoding space roughly follow a unit Gaussian distribution. 
The decoder learns to reconstruct the datapoint \(x\) given the representation \(z\), denoted by \(p_{\phi}(x|z) \). 
VAE maximizes $\mathbb{E}_{z \sim q_{\theta}(z|x)}[\log p_{\phi}(x|z)] - \text{KL}(q_{\theta}(z|x) || p(z))$, where $p(z)$ is the unit Gaussian distribution. 
The first term is the reconstruction likelihood, which encourages the decoder to learn to reconstruct $x$. 
The second term is KL-divergence that ensures \(q_{\theta}(z|x)\) is similar to the prior distribution \(p(z)\). 
This has the effect of keeping the representations of similar datapoints close together rather than separated in different regions of the encoding space.

\subsubsection{Kernel Density Estimation} 
KDE belongs to the class of non-parametric density estimations. 
Closely related to histograms, but KDE smooths out the contribution of each observed datapoint \(x_i\) over a local neighborhood of that datapoint by centering a kernel function. Formally, KDE can be formulated as 
\[ \hat{f}_h(x)=\frac{1}{nh}\sum_{i=1}^{n}\mathsf{K}(\frac{x-x_i}{h}),\]
where \(\mathsf{K}\) is the kernel function, and \(h>0\) is the bandwidth that controls the amount of smoothness. 
Due to the convenient mathematical properties, the Gaussian kernel is often used. 
The choice of bandwidth is a tradeoff between the bias of estimator and its variance. 
 	
\begin{figure}[t]
 		\centering
 		\includegraphics[width=.4\textwidth]{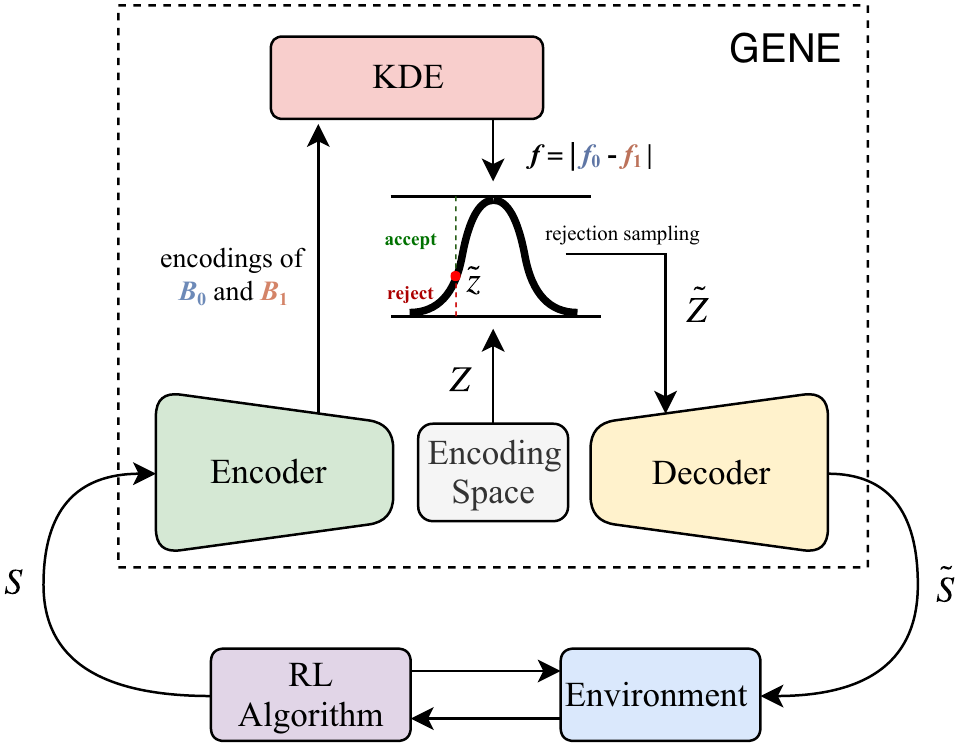}
 		\caption{GENE consists of a VAE and a KDE. Samples from the encoding space of experienced states are passed through rejection sampling and then fed into the decoder to generate start states.}
 		\label{fig:gene}
\end{figure}

\section{Method}

When we humans learn to solve a task, we never always start from the very beginning, but stand up from where we fall down and move forward. 
More specifically, we deliberately practice more on some \emph{unfamiliar} and \emph{unskilled} states. 

The basic idea of GENE follows this intuition. 
At the beginning, the agent is not able to reach the goal and hence GENE generates start states with low density in the distribution of states experienced by the agent. 
Low density means the generated states are \emph{novel} states (\emph{i.e.}, the agent is unfamiliar with), and starting from these states the agent is able to explore the environment further. 
When novel states become common (\emph{i.e.}, higher density than before), new novel states will be generated. 
Therefore, GENE propels the agent to explore the environment gradually. 
The aim of exploration is to obtain reward signals. 
After the agent obtains the reward signal, there exist some experienced states from which the current learned policy is only possible to reach the goal. We call them \emph{unskilled} states (\emph{i.e.}, the agent is unskilled at). 
Thus, the agent needs more training on these unskilled states. 
As the policy improves and the agent masters the previous unskilled states, new unskilled states are continuously generated by GENE and gradually trace back to the initial state until the task is solved. 
In short, GENE guides the agent to explore the environment by starting from the novel states and reinforces the learned policy by starting from the reversing unskilled states.

\subsection{State Generation}

GENE consists of a VAE and a KDE and works with any RL algorithm, as illustrated in Figure~\ref{fig:gene}. 
In a training episode, if the agent does not reach the goal, we store all the states experienced in this episode, called failed states, in the buffer \(\mathcal{B}_0\), otherwise we store the states, called successful states, in another buffer \(\mathcal{B}_1\). 
It is obvious that the agent starting from the states in \(\mathcal{B}_1\) will be more likely to reach the goal than starting from the states in \(\mathcal{B}_0\).

In order to purposely generate novel states and unskilled states, it is necessary to estimate the state distributions of $\mathcal{B}_0$ and $\mathcal{B}_1$. 
However, the density estimation of high-dimensional states is usually intractable. 
Fortunately, the encoder of VAE maps the high-dimensional state to the encoding space which is described as \(k\)-dimension mean and log-variance \((\mu,\log\sigma)\). 
We use the mean value $\mu$ as the encoding of the input state. 
As the encoding space is only $k$-dimension and roughly follows the unit Gaussian distribution, it is easy to estimate the PDFs of the encodings of the states in $\mathcal{B}_0$ and $\mathcal{B}_1$, denoted by $f_0$ and $f_1$ respectively. 
We use KDE as the PDF estimator. 
It produces a more smooth PDF based on individual locations of all sample data without suffering from data binning, which makes it more suitable for the continuous variable.

We uniformly sample from the encoding space to get a set of encodings $Z$. 
Then rejection sampling is applied to select eligible encodings from $Z$. 
The principle is to give a high probability to the encoding with low $f=|f_0-f_1|$. 
We propose a uniform distribution with the PDF \( (1+\epsilon)*\max(f)\). 
Every time we randomly take out an encoding \(\tilde{z}\) from $Z$ and sample a random number \(u\) from \(\mathrm{Unif}(0, (1+\epsilon)*\max(f))\).  
If \(f(\tilde{z})<u\),  we accept \(\tilde{z}\), otherwise we reject it, as illustrated Figure~\ref{fig:gene}. 
Repeat the sampling process until the number of accepted samples $\tilde{Z}$ is equal to \(T\), which is a training parameter and will be discussed in the following. 
Then, pass \(\tilde{Z}\) to the decoder to reconstruct the states $\tilde{S}$, from which the agent will start new episodes.
 	
\begin{figure}[t!]
	\centering
	\includegraphics[width=.38\textwidth]{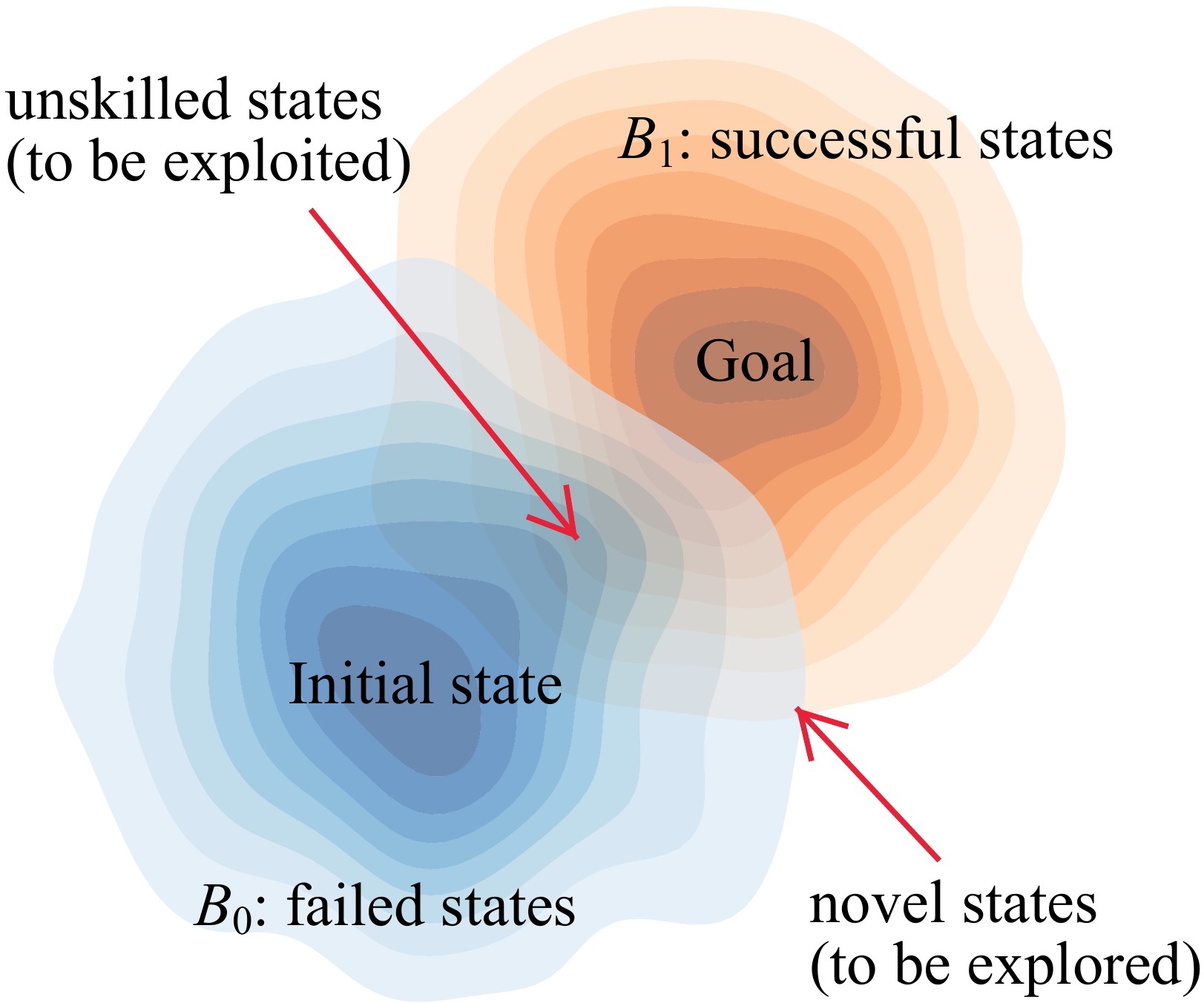}
	\caption{Illustrating the mechanism of GENE.}
	\label{fig:demo}
\end{figure}
 	 	
The mechanism of GENE is illustrated in Figure~\ref{fig:demo}. At the beginning, since the agent is not able to reach the goal, $\mathcal{B}_1$ is empty and hence $f_1=0$. 
$\mathcal{B}_0$ contains all the states the agent has recently experienced, and $f=f_0$. Thus, $f$ is currently the density of recently experienced states. 
Therefore, the generated states with low $f$ are novel states, and starting from these states could help exploration. 
When novel states become common, new novel states will be generated for further exploration. 
When there are successful states in $\mathcal{B}_1$ (\emph{i.e.}, the agent has reached the goal at least once), GENE will generate states according to $f=|f_0-f_1|$. 
Since the current policy is possible to reach the goal but still requires more training when starting from unskilled states, the unskilled states are with low $|f_0-f_1|$ and more likely to be generated. .
Also there are some states with low densities in both $\mathcal{B}_0$ and $\mathcal{B}_1$, which are also likely to be generated and worth exploring.

Generally, VAE tends to generate data with noise, which is an obvious shortcoming in computer vision (\emph{e.g.}, blurry images). 
However, in our case, the generated states with noise actually prevent the agent from always repeating the states it has experienced and thus help the exploration, making GENE more sample-efficient. 
As the policy updates, the two distributions of experienced states also vary, which brings two benefits. 
On the one hand, novel states become common gradually, which propels the agent to explore new novel states continuously. 
On the other hand, unskilled states are generated gradually from near the goal to near the initial state without any prior knowledge. Thus, GENE can automatically tradeoff between exploration and exploitation to guide the policy optimization. 
We will further investigate this in the experiments.

\subsection{Training}

Algorithm~\ref{alg:1} details the training of GENE. 
Every episode, the agent starts from the generated states $\tilde{S}$ with a probability $p$, otherwise from the initial state. 
The probability $p$ could be seen as how much to change the start state distribution. 
If it is too small, the effect is insignificant, and if it is too large, the agent cannot focus on the original task (from initial state). 
Ablation studies in the next section will show how the probability $p$ affects the performance. 
Every \(T\) episodes, we train the VAE from the scratch using the states stored in \(\mathcal{B}_0\) and \(\mathcal{B}_1\). 
Training from the scratch every \(T\) episodes helps avoid overfitting and collapse when the distribution of experienced states changes slowly. 
Training VAE is efficient and stable and would not be a bottleneck.
The PDFs of the experienced states are estimated and fitted by KDE via their encodings. 
Then, $\tilde{Z}$ is obtained by applying rejection sampling to $Z$, and the states are generated by the decoder for the next \(T\) episodes. The RL model is updated at the end of every episode, which is independent of the state generation. As GENE does not directly interact with the RL algorithm, it is very easy to implement and compatible with any RL algorithm, no matter on-policy or off-policy, single-agent or multi-agent.

 	\begin{algorithm}[t]
 		\caption{Generative Exploration and Exploitation}
 		\label{alg:1}
 		\begin{algorithmic}[1]
 			\STATE Initialize an RL model (\emph{e.g.}, PPO, TRPO, DDPG)
 			\STATE Initialize state buffers \(\mathcal{B}_0\) and  \(\mathcal{B}_1\)
 			\FOR{episode = $1, \ldots, \mathcal{M}$}
 			\STATE Store failed states in \(\mathcal{B}_0\) 
 			\STATE Store successful states in \(\mathcal{B}_1\) 
 			\IF{ \(\text{episode} \% T = 0\) }
 			\STATE Train a VAE using \(\mathcal{B}_0+\mathcal{B}_1\) 
 			\STATE Fit $f_0$ of $\mathcal{B}_0$ and $f_1$ of $\mathcal{B}_1$ using the encodings via KDE
 			\STATE Sample from the encoding space to obtain $Z$ 
 			\STATE Apply rejection sampling to select \(\tilde{Z}\) from $Z$ according to $|f_0-f_1|$
 			\STATE Reconstruct states $\tilde{S}$ from $\tilde{Z}$ for next \(T\) episodes 
 			\STATE Clear the buffers \(\mathcal{B}_0\) and  \(\mathcal{B}_1\)
 			\ENDIF
 			\STATE Update the RL model
 			\STATE The agent starts from generated states $\tilde{S}$ in a certain probability $p$
 			\ENDFOR
 		\end{algorithmic}
 	\end{algorithm}

 	\begin{figure*}[t]
 		\centering
 		\begin{subfigure}[t]{0.23\textwidth}
 			\centering
 			\includegraphics[height=1\textwidth]{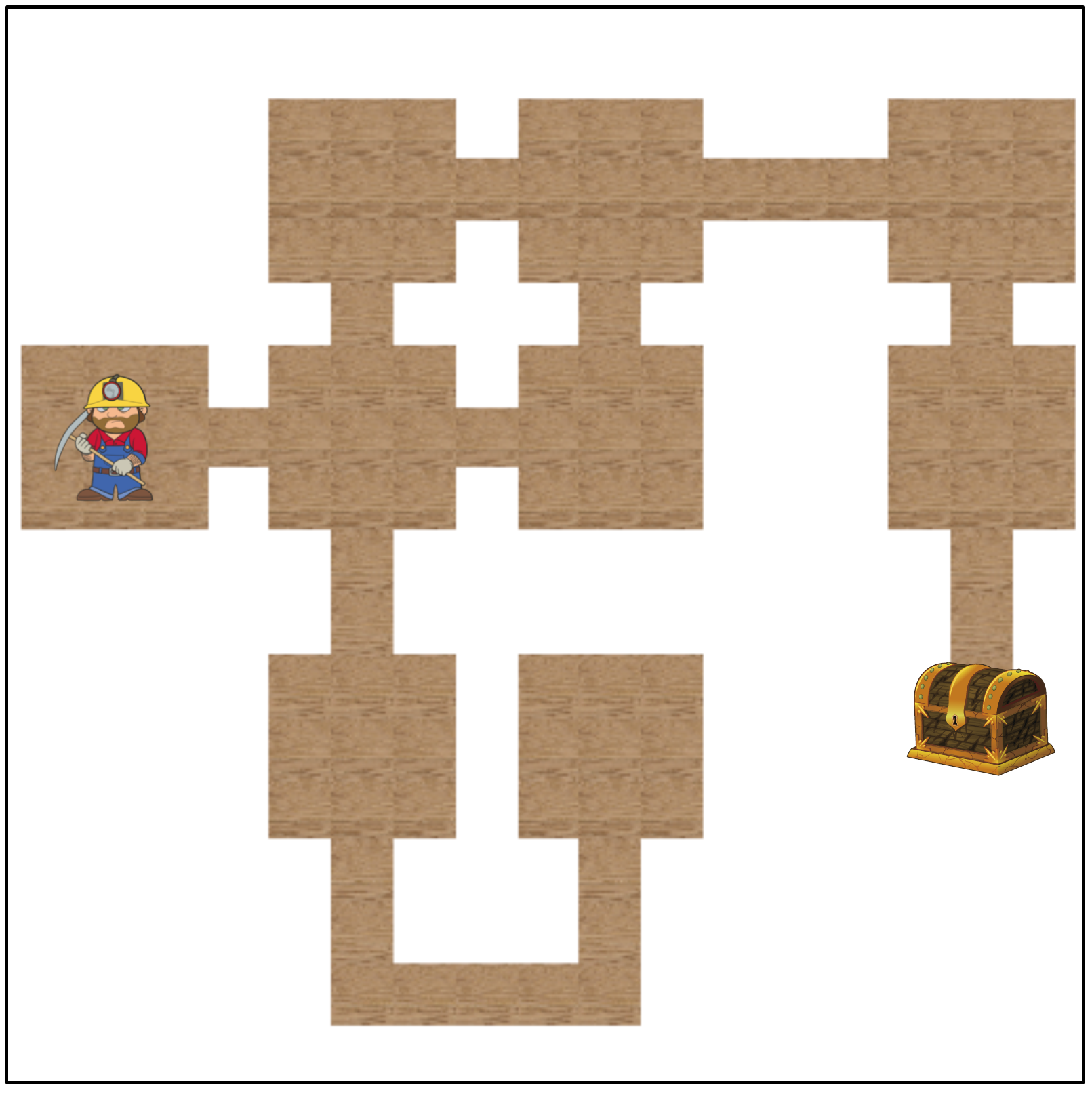}
 			\caption{Maze}
 			\label{fig:maze}
 		\end{subfigure}
 		\hspace{0.1cm}
 		\begin{subfigure}[t]{0.23\textwidth}
 			\centering
 			\includegraphics[height=0.995\textwidth]{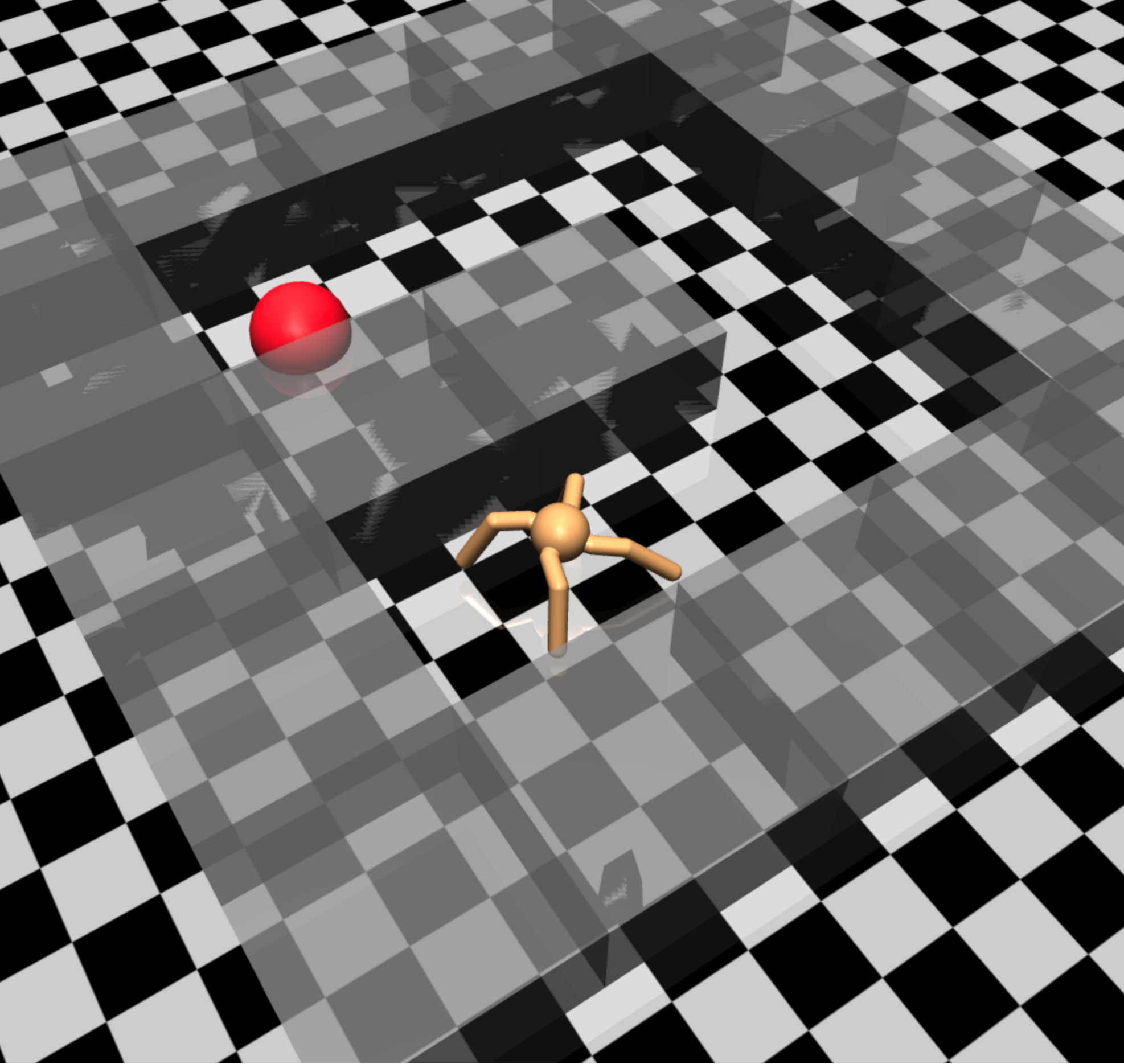}
 			\caption{Maze Ant}
 			\label{fig:maze_ant}
 		\end{subfigure}
 		\hspace{0.2cm}
 		\begin{subfigure}[t]{0.23\textwidth}
 			\centering
 			\includegraphics[height=1\textwidth]{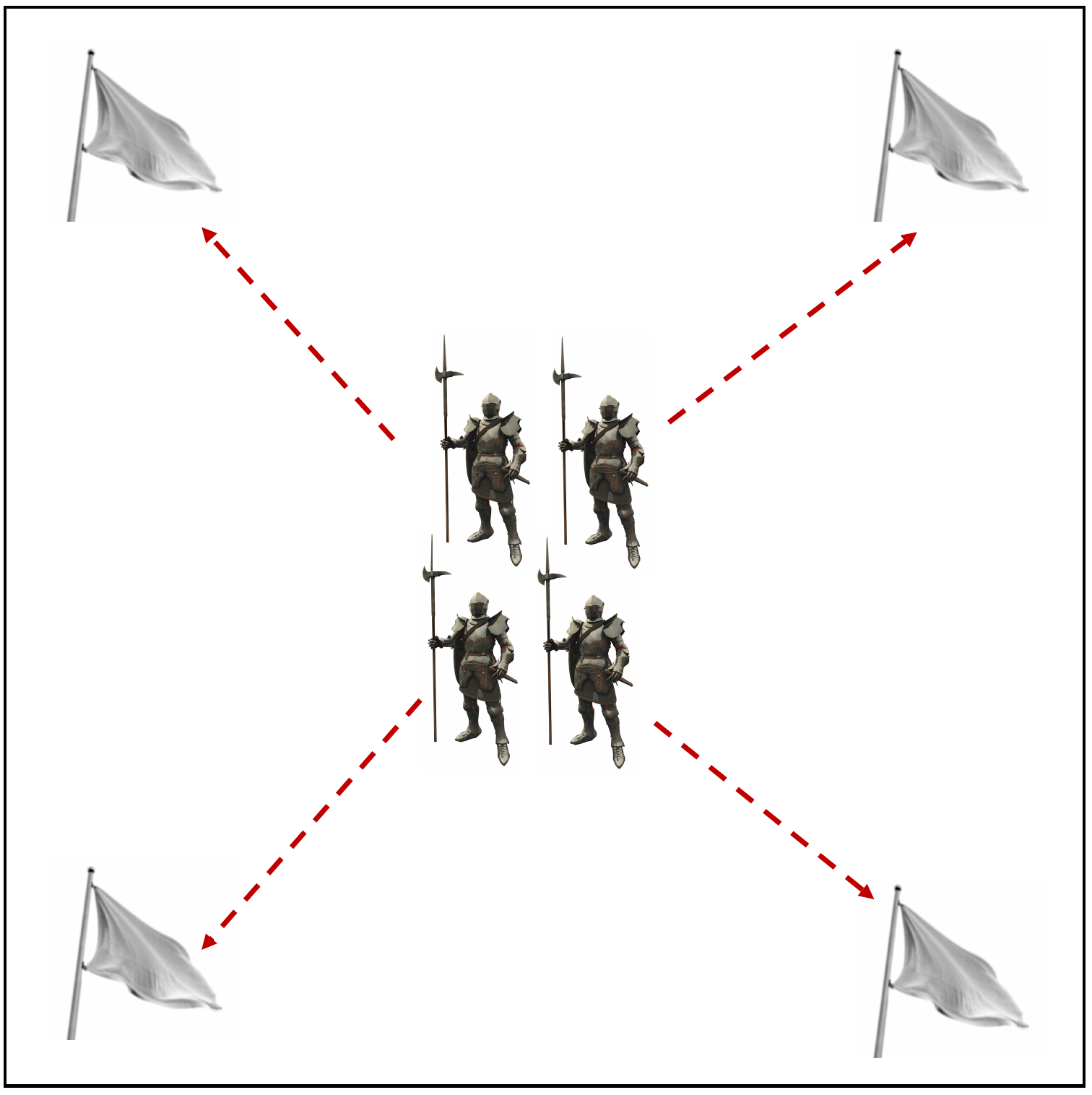}
 			\caption{Cooperative Navigation}
 			\label{fig:cn}
 		\end{subfigure}
 		\vspace{-0.15cm}
 		\caption{Illustrations of experimental tasks with binary rewards.}
 		\label{fig:scneraios}
 		\vspace*{-0.1cm}
 	\end{figure*}

\section{Experiments}
	 	
In this section, we focus on the following questions:
 	\begin{itemize}
 		\item Can the mechanism and effectiveness of GENE be verified and interpreted by experiments?
 		\item Is GENE effective and efficient in high-dimensional environments?
 		\item Is GENE suitable in multi-agent environments?
 	\end{itemize}

To answer these questions, we investigate GENE in three tasks with binary rewards indicating whether or not the task is completed. 
To verify the exploration effectiveness, we compare GENE with three popular exploration methods, \textit{RND} \cite{burda2018exploration} that quantifies state novelty as intrinsic reward), \textit{Goal GAN} \cite{florensa2018automatic} and \textit{HER} \cite{andrychowicz2017hindsight}  that set additional goals. 
As for the reversing effect, we compare it against four methods that change the start state distribution.
 	\begin{itemize}
 		\item \textit{Uniform}, sampling start states from the uniform distribution and thus assuming prior knowledge about the environment.
 		\item \textit{History}, sampling start states from the agent's historical states.
 		\item \textit{Demonstration} \cite{nair2018overcoming,resnick2018backplay}, assuming access to the successful demonstration and sampling start states from demonstration states.
 		\item \textit{RCG} \cite{florensa2017reverse}, setting start states which are between the bounds on the success probability \([R_\text{min},R_\text{max}]\) by taking random walks from the goal state.
 	\end{itemize}

Both GENE and the baselines work on a base RL algorithm. 
The parameters of the base RL algorithm are the same, which guarantees the comparison fairness. 
To answer the first question, we demonstrate GENE in a challenging Maze (Figure~\ref{fig:maze}).
For the second question, we study GENE in a robotic locomotion tasks, Maze Ant (Figure~\ref{fig:maze_ant}). 
For the last question, we demonstrate GENE in Cooperative Navigation (Figure~\ref{fig:cn}), a typical multi-agent cooperative task. 
The details of each task and the hyperparameters of the algorithms used in the experiments are available in Appendix.
All the experimental results are presented using mean and standard deviation of five runs. 

\subsection{Maze}
 	
In the 2D maze, the agent learns to navigate from an initial position to the target position within a given number of timesteps as depicted in Figure~\ref{fig:maze}. 
Only if the agent reaches the target, it receives a reward \(+1\). 
In Maze, we choose PPO \cite{schulman2017proximal} as the base RL algorithm.
 	
 	 \begin{figure}[b!]
 		\centering
 		\includegraphics[width=0.37\textwidth]{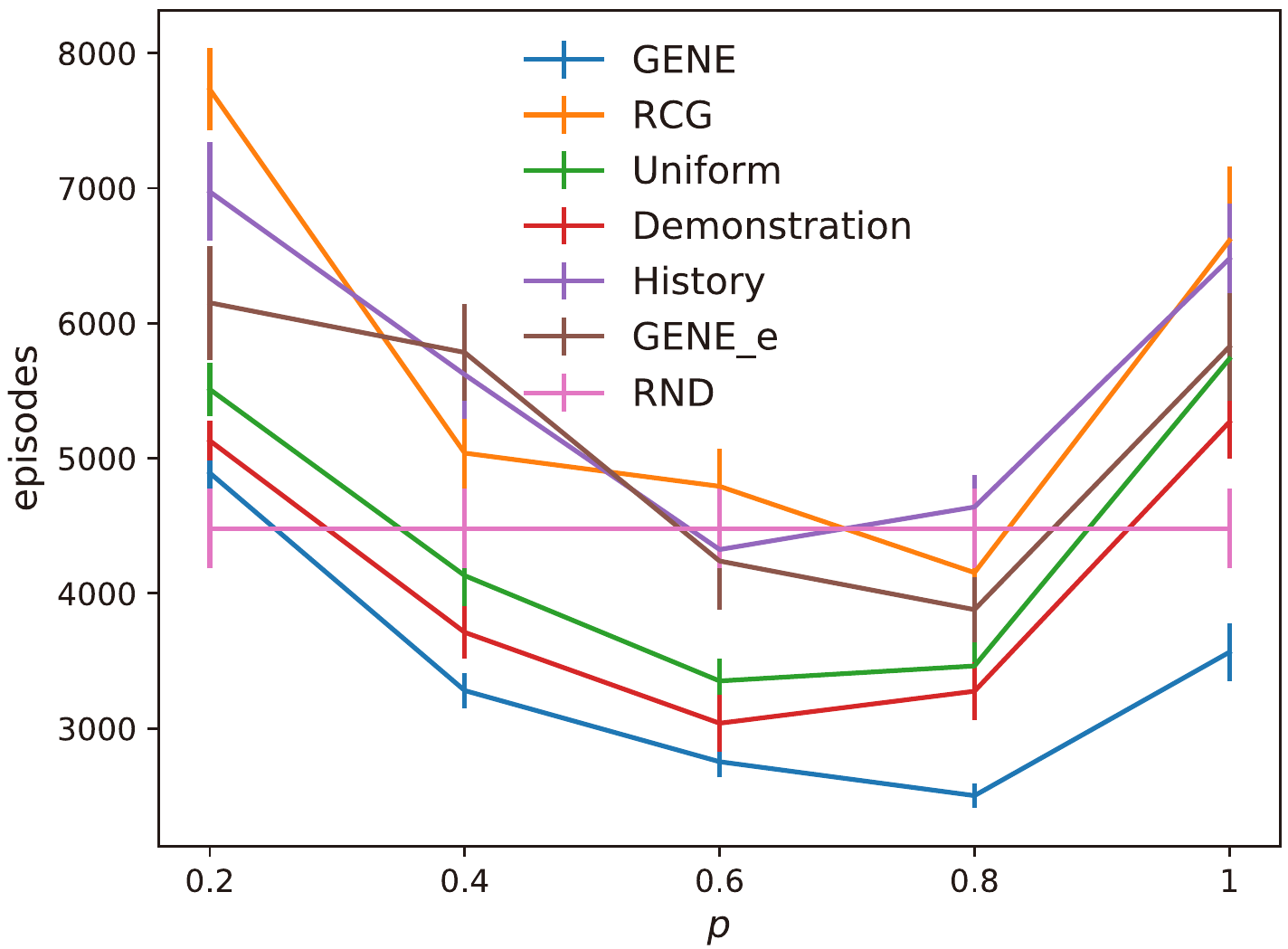}
 		\vspace{-0.15cm}
 		\caption{Episodes to solve the task with different probabilities $p$.}
 		\label{fig:maze_curve}
 	\end{figure}
 	
Figure~\ref{fig:maze_curve} shows the number of episodes to solve the task (\emph{i.e.}, achieving ten consecutive successes starting from the initial state) with different $p$ of changing start state distribution. When $p=0$, the algorithm degenerates into the base algorithm PPO, which suffers from prohibitive amount of undirected exploration to reach the goal and is incapable of solving this task. 
When $p$ is too small, the effect of changing start state distribution is insignificant. While the $p$ is around $1.0$, the agent does not get enough training on the initial position, as a result it takes more episodes to solve the original task.

GENE agent learns more quickly than other baselines, which is attributed to that it focuses on the novel states and unskilled states and adaptively tradeoffs between them. 
Uniform agent spends many episodes on the useless area, such as the dead end at the bottom of the maze. 
Sampling from the demonstration could avoid exploring the useless area, but uniformly sampling from the demonstration cannot make the agent focus on the instructive states. So both methods spend more episodes than GENE. 
Sampling from the agent's history requires no prior knowledge, but it gives higher probability to more familiar states, which however could be easily visited and unworthy of practice. 
Therefore, it barely helps. 
Although RCG automatically generates start states in reverse, growing outwards from the goal. 
It assumes access to the goal state, a priori knowledge, which means RCG ignores the exploration progress. 
Moreover, RCG requires to test whether the success probability of candidate states between the bounds on the success probability \([R_\text{min},R_\text{max}]\). 
This incurs much more additional episodes. In addition, \(R_\text{min}\) and \(R_\text{max}\) are manually tuned hyperparameters, which can greatly affect the overall performance and requires careful tuning.

To verify the exploration effectiveness of GENE, we compare it against RND \cite{burda2018exploration}. 
In GENE, the generated novel states encourage the agent to explore. 
From Figure~\ref{fig:maze_curve}, we can see that GENE takes less episodes than RND when $p\geq 0.4$. The shaped reward of RND is biased from the true target, \emph{e.g.}, leading the agent to the dead end, which causes much more episodes. 
For further investigation, we make $f=f_0$, \emph{i.e.}, to only generate novel states, termed GENE\_e. 
GENE\_e still outperforms RND when $p=0.6 \text{ and }  0.8$, which demonstrates just starting from novel states could better help exploration. 
The difference between GENE and GENE\_e verifies that replaying unskilled states truly accelerates the learning.

\begin{figure*}[t!]
\centering
\begin{minipage}{0.6\textwidth}
\centering
\includegraphics[width=1\textwidth]{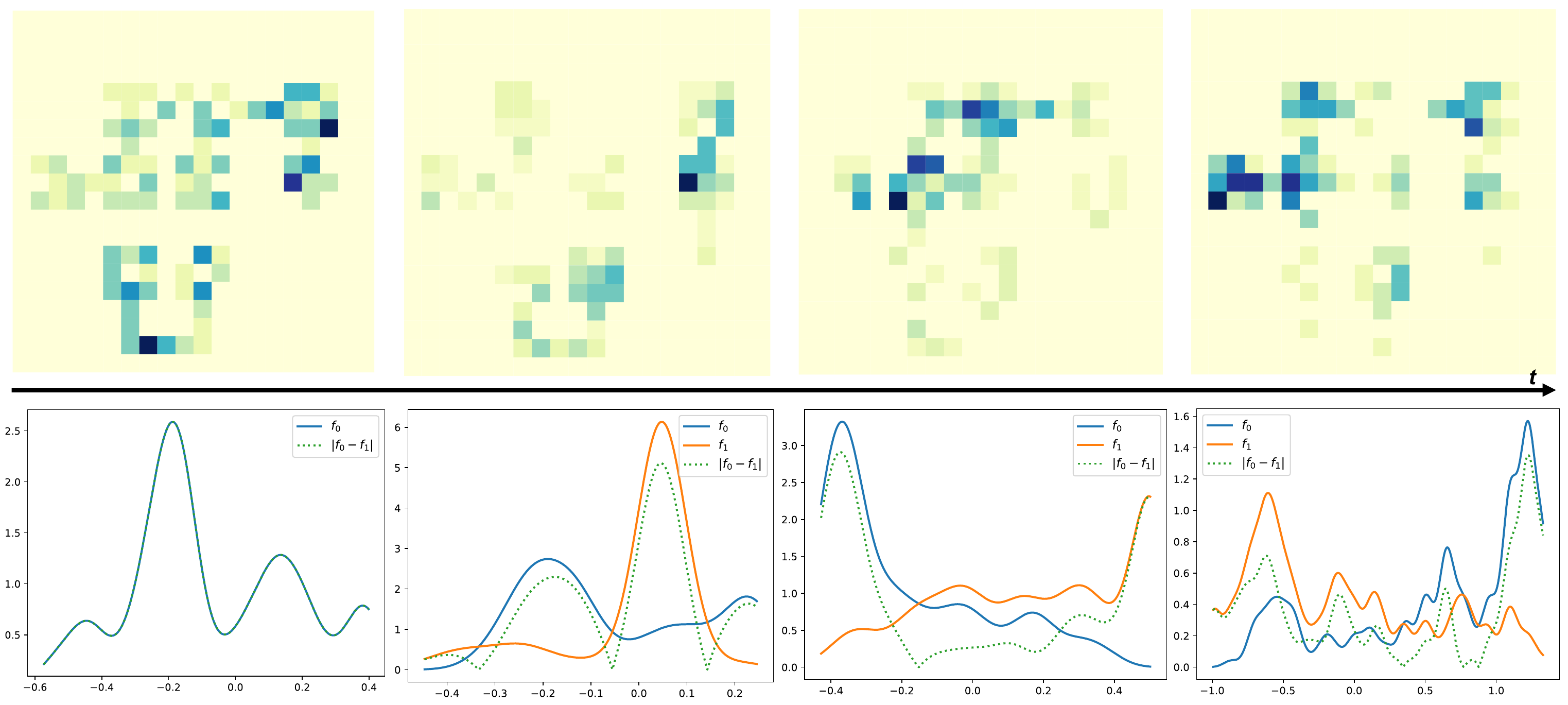}
\vspace{-0.6cm}
\caption{Top row shows the heatmaps of generated states as the training progresses. Bottom row shows the PDFs over the encoding space, where $f_0$ corresponds to the blue, $f_1$ corresponds to the orange, and $f$ corresponds to the green.}
\label{fig:maze_process}
\end{minipage}
\hspace{0.25cm}
\begin{minipage}{0.37\textwidth}
\vspace{-0.5cm}
\centering
\includegraphics[width=1\textwidth]{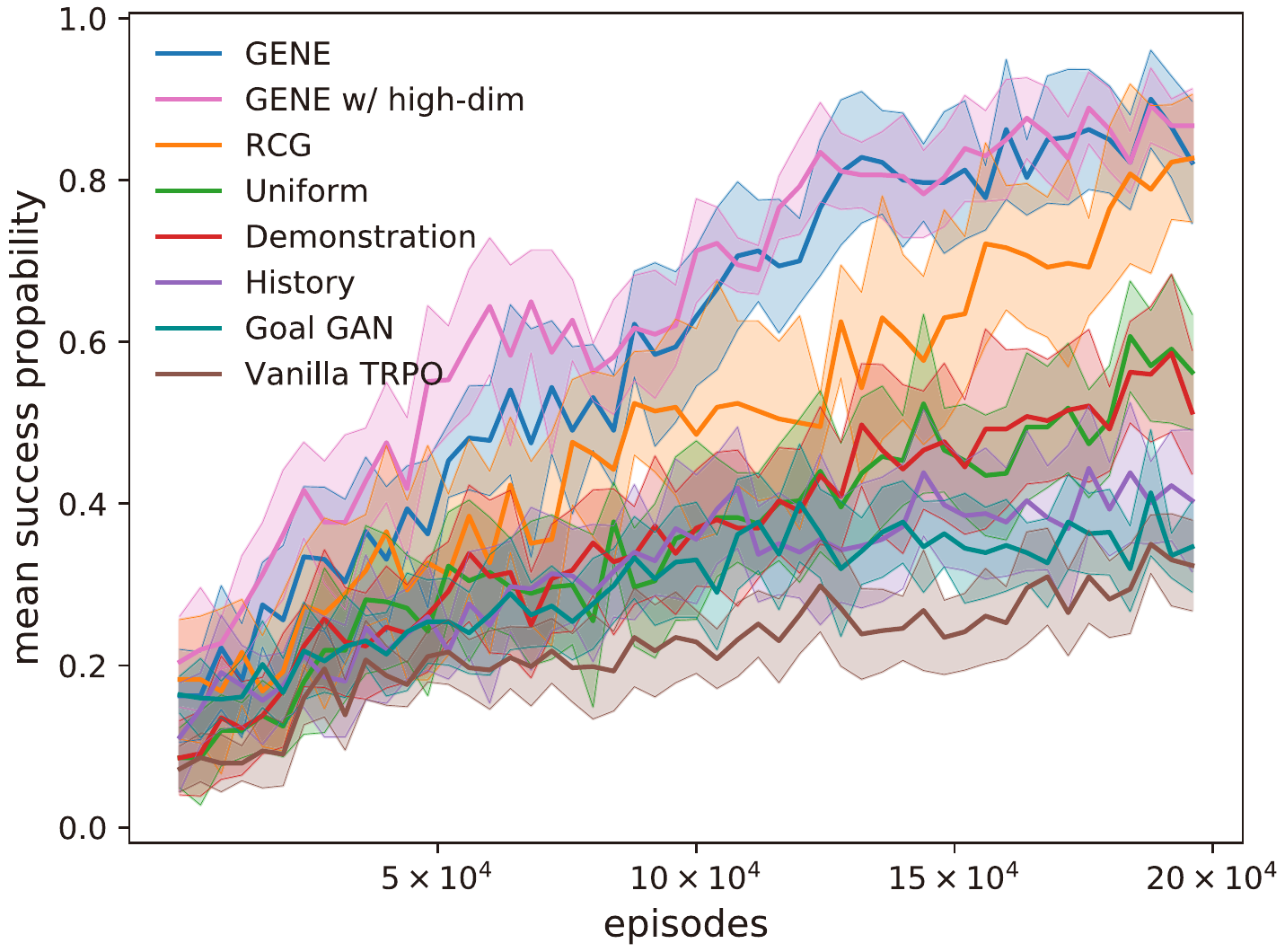}
\vspace{-0.5cm}
\caption{Learning curves in Maze Ant.}
\label{fig:ant_curve}
\end{minipage}
\vspace*{-0.1cm}
\end{figure*}

Figure~\ref{fig:maze_process} gives more details of the learning process and explains the mechanism of GENE. 
At the beginning, $\mathcal{B}_1$ is empty and $f=f_0$. 
By giving high probability to states with low $f_0$, novel states are generated. 
The agent is wandering around the start position, so the generated states are mostly distributed at the edge of the activity scope. 
As the training progresses, the agent becomes familiar with the states which are originally novel and the agent's activity scope gradually expands. 
Subsequently, the agent can reach the goal occasionally, then there are successful states stored in $\mathcal{B}_1$.
States with low $|f_0-f_1|$ are possible for the current policy to reach the goal, but the agent still requires more training. 
Moreover, as illustrated in Figure~\ref{fig:maze_process} (top row), in the generated states the distance between the agent and the goal gradually increases. 
This is because as the policy improves, the early unskilled states are easy for the agent and thus more difficult states are generated. 
The learned policy is continuously optimized by the generated states with gradually increased difficulty. 
This is an obvious reversing effect. 
When the generated states trace back to the initial state, the task is solved and there is no need to pay attention to the dead end at the bottom of the maze. This makes GENE more efficient. 

\subsection{Maze Ant}
 	
The ant within a U-shaped maze tries to reach the goal from a fixed initial position within a given number of timesteps, as illustrated in Figure~\ref{fig:maze_ant}. 
Only when the ant gets the goal, it receives a reward \(+1\). 
The state space is 37-dimension, including the positions of the ant and the positions and velocities of the ant's joints. 
The action space of the ant is 8-dimensional, controlling the movement. 
In Maze Ant, we choose TRPO \cite{schulman2015trust} as the base RL algorithm.

\begin{figure}[b!]
 		\centering
 		\includegraphics[width=0.48\textwidth]{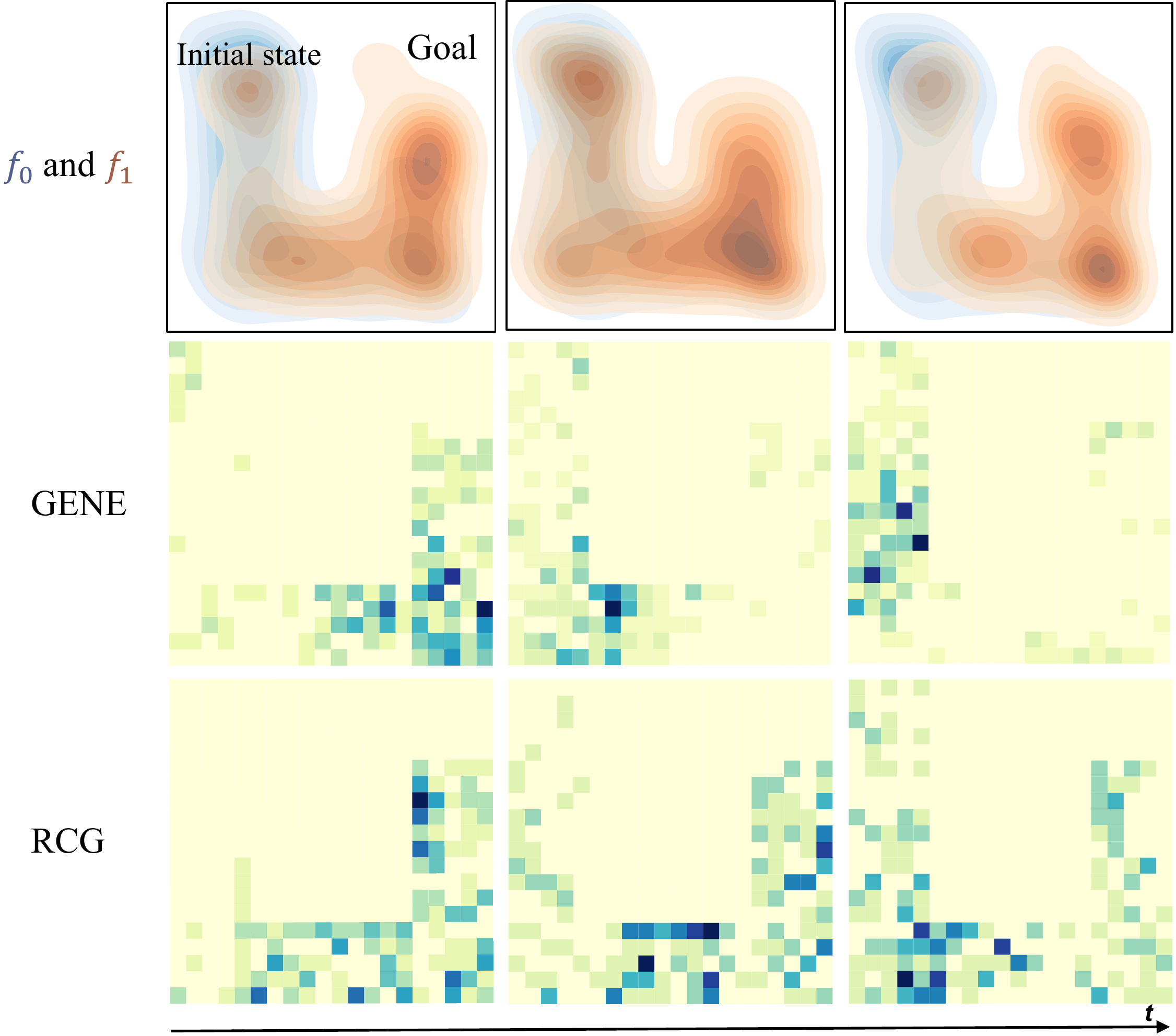}
 		\caption{Visualizations of $f_0$ (blue) and $f_1$ (orange) of GENE, and the heatmaps of GENE and RCG in three different training episodes.}
 		\label{fig:ant_distribution}
 	\end{figure}

Figure~\ref{fig:ant_curve} shows the learning curves of GENE and the baselines. 
Vanilla TRPO is in trouble with learning in this sparse reward environment. 
As there is only one way from the initial position to the goal, the performance of Uniform and Demonstration is similar. 
GENE outperforms RCG because the generated states of GENE are more focused and reverse more quickly than RCG's random walk, which is well illustrated in Figure~\ref{fig:ant_distribution}. 
That shows the states generated by GENE are more helpful. 
From the visualizations of $f_0$ and $f_1$ and the heatmaps of GENE, we can see that the generated states are mainly distributed in the regions where $f_0$ and $f_1$ balance and trace back automatically as $f_0$ and $f_1$ change. 
As illustrated in Figure~\ref{fig:ant_distribution}, at the early stage, only starting from the states closed to the goal the agent is likely to reach the goal, so there is a peak of $f_1$ near the goal. 
As the policy improves, the $f_1$ peak traces back, and correspondingly the generated states move farther away from the goal.
Gradually, there are several $f_1$ peaks along the path, meaning the agent has mastered most states in the maze, and the generated states are mostly located near the initial state.
 	
To investigate whether changing start states is more efficient than setting additional goals in single-goal situations, we compare GENE against Goal GAN. 
The training set of Goal GAN is uniformly sampled from the goal space and we evaluate the performance on the target goal. 
Figure~\ref{fig:ant_curve} shows GENE substantially outperforms Goal GAN. 
Before overcoming the target goal, Goal GAN must master a serial of easy goals, which distracts the policy and increases the training difficulty.
 	
 	 \begin{figure}[t!]
 		\centering
 		\includegraphics[width=0.37\textwidth]{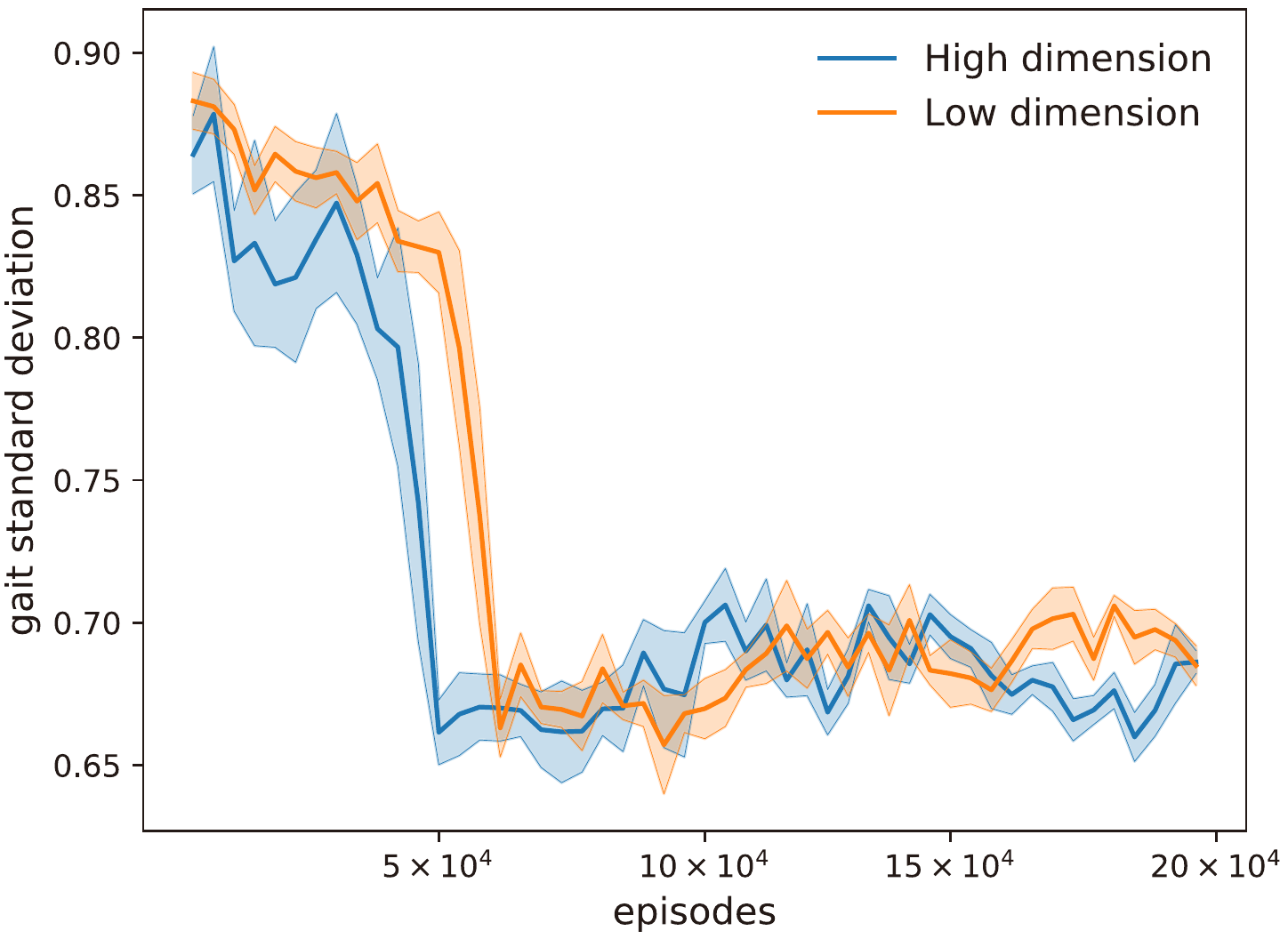}
 		\vspace{-0.15cm}
 		\caption{Standard deviation of gait in Maze Ant.}
 		\label{fig:std_curve}
 	\end{figure}

Only 2-dimensional positions of the ant are generated in the experiments above. 
To investigate whether GENE could deal with complex state with high dimension, we apply GENE to generate the positions and velocities of the ant's joints with totally 37 dimensions, termed GENE w/ high-dim. The control of multi-joint robot is complex due to the high degrees of freedom and issues such as gimbal lock. 
The success of GENE w/ high-dim explains the generativity in high-dimensional state space, which is attributed to that VAE could map the high-dimensional state to a meaningful encoding space. 
This also helps the learning. 
To reach the goal, the ant must learn how to crawl first. 
GENE w/ high-dim generates adequate postures for the ant to explore how to crawl, so the ant learns to crawl more quickly than GENE as illustrated by the curves of the standard deviation of the ant gait (the ant torso, \emph{e.g.}, the joints' positions and orientations, in an episode) in Figure~\ref{fig:std_curve}. 
When the ant masters how to crawl, the gait is more steady and hence the standard deviation decreases. 
Benefited from this, GENE w/ high-dim learns more quickly than GENE in the early stage as depicted in Figure~\ref{fig:ant_curve}.

Table~\ref{tab:proportion} gives the proportion in training time of GENE in Maze Ant. 
We can see training VAE only takes 11\%. Thus, the training of VAE is efficient and would not be a bottleneck. 
Also it is known that the distribution of VAE's outputs obeys the distribution of the training set, thus the probability of generating unreasonable states is low. 
According to statistical result, there are only 2.8\% unreasonable states, \emph{e.g.}, the ant is not located in the maze field. 
However, these states can be easily refused by the simulator without affecting the performance.
 	
 	\begin{table}[h]
 		\centering
 		\caption{Proportion in training time of GENE in Maze Ant}
 		\label{tab:proportion}
 		\vskip -0.2cm
 		\begin{footnotesize}
 			\begin{tabular}{@{}ccc@{}}		
 				\toprule
 				Interaction   & Training TRPO & Training VAE     \\ 
 				\midrule
 				$74\%$     &   $15\%$  &   $11\%$\\ 
 				\bottomrule
 			\end{tabular}
 		\end{footnotesize}
 	\end{table}

\subsection{Cooperative Navigation}
 	
In multi-agent environments, many tasks rely on collaboration among agents. 
However, the agent does not know the policies of others and their policies are always changing during training, and thus the task is much more difficult than the single-agent version. 
In this Cooperative Navigation task, there are a same number of landmarks and agents. 
The goal of agents is to occupy each landmark within a given number of timesteps, as illustrated in Figure~\ref{fig:cn}. 
Only when every landmark is occupied by an agent, each agent receives a reward \(+1\). 
Therefore, this is a case of binary reward for the multi-agent environment. 
We choose MADDPG \cite{lowe2017multi} as the base multi-agent RL algorithm, where each agent has an independent actor-critic network, without weight sharing or communication. 
 	
 	\begin{figure}[t]
 		\centering
 		\includegraphics[width=0.37\textwidth]{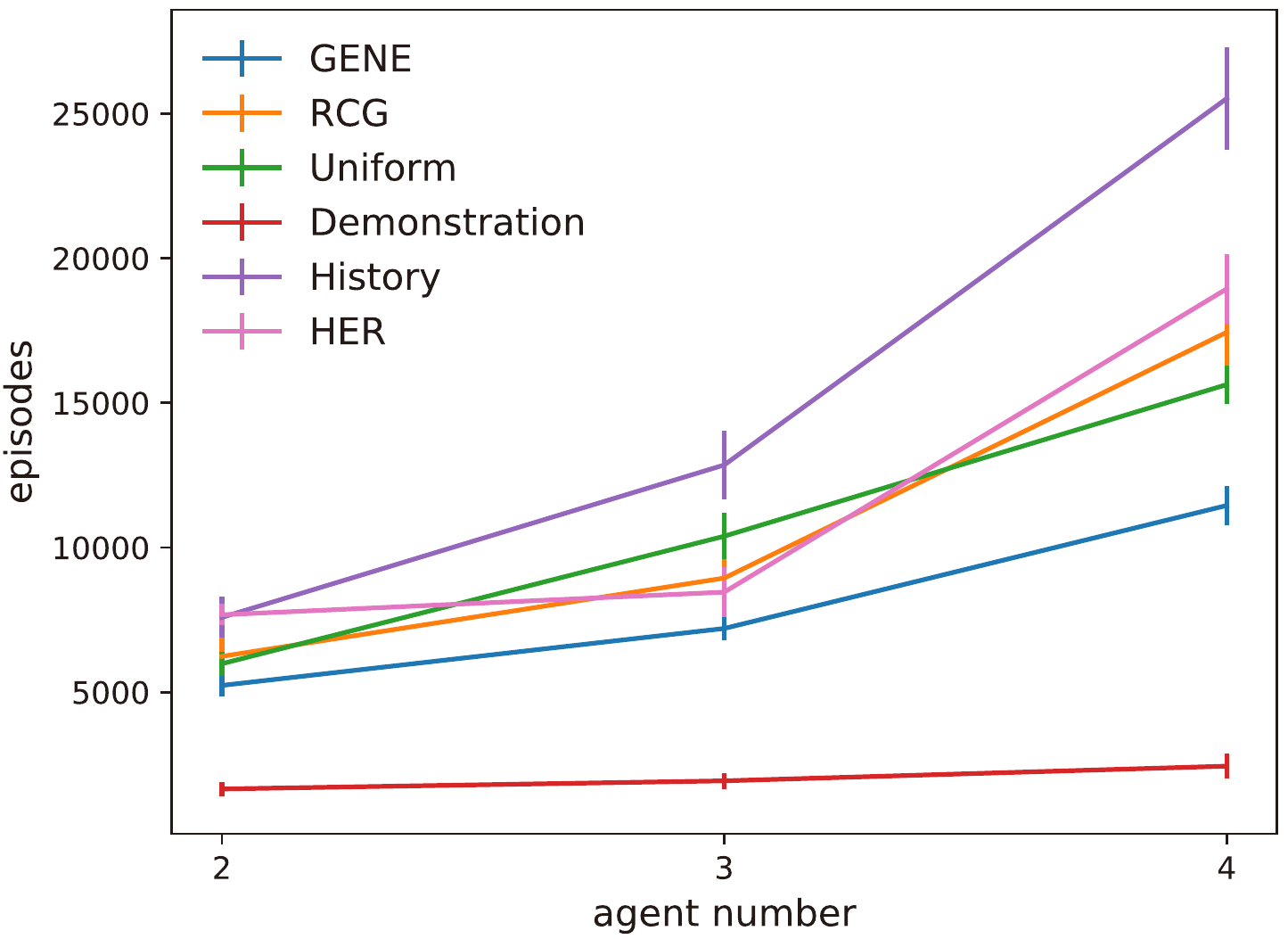}
 		 		\vspace{-0.15cm}
 		\caption{Episodes to solve Cooperative Navigation with different agent numbers.}
 		\label{fig:cn_curve}
 	\end{figure}
 	
Figure~\ref{fig:cn_curve} shows the training episodes to solve Cooperative Navigation with different number of agents. 
Vanilla MADDPG cannot solve this task, because the agents hardly occupy the landmarks simultaneously with random exploration, \emph{e.g.}, Ornstein-Uhlenbeck noise. 
Demonstration agents spend the least episodes, because the experience in the successful demonstration dramatically reduces the difficulty of the task. As each agent only samples the states from corresponding agent, the agent number does not impact its performance much. 
However, note that \textit{obtaining the successful demonstration itself is very challenging in this task}. 
RCG's random walk from the goal state progresses very haphazardly in such an open field. 
The agents do not know which landmark to cover in advance and must learn the division of roles. 
Uniformly sampling would cause two agents cover the same landmark, which yields no reward signals and does not help for division of roles. 
GENE makes the agents practice more on the states from which there is a certain probability to cover all the landmarks, and thus encourages the agent to learn its own role. 
When the number of agents increases, the search space increases exponentially and it becomes less possible that every landmark is occupied at the same time, thus the reward is extremely sparse. 
However, the gain of GENE over other baselines even expands with the increase of agents. 
This indicates GENE indeed accelerates the learning of multiple agents in the cooperative task regardless the number of agents. 
 	
To verify the ability of exploration in this task, we apply HER to MADDPG as a baseline of exploration method. HER is proposed for DDPG but exactly matches MADDPG. As depicted in Figure~\ref{fig:cn_curve}, GENE outperforms HER. 
Although setting arbitrary experienced states as additional goals could help exploration, HER agents have to learn many additional goals and rarely obtain a real reward signal, which slows down the learning. 

\section{Conclusions}
In this paper, we have proposed GENE for overcoming sparse rewards in RL. 
By dynamically changing the start state of agent to the generated state, GENE can automatically tradeoff between exploration and exploitation to optimize the policy as the learning progresses. 
GENE relies on no prior knowledge about the environment and can be combined with any RL algorithm, no matter on-policy or off-policy, single-agent or multi-agent. Empirically, we demonstrate that GENE substantially outperforms existing methods in a variety of tasks with binary rewards.

\section{Acknowledgments}
This work was supported by NSFC under grant 61872009.

\bibliographystyle{aaai}
\bibliography{reference}
 	

 	\appendix
 	
 	\begin{table*}[b!]
 		\renewcommand{\arraystretch}{1.1}
 		\centering
 		\caption{Hyperparameters}
 		\label{tab:hyperparameter}
 		\vskip -0.2cm
 		\begin{footnotesize}
 			\begin{tabular}{@{}cccc@{}}
 				\toprule
 				Hyperparameter&Maze&Maze Ant&Cooperative Navigation\\
 				\midrule
 				maximum timesteps &$50$&$200$&$50$\\
 				RL algorithm&PPO&TRPO&MADDPG\\
 				discount (\(\gamma\))&$0.98$&$0.99$&$0.95$\\
 				batch size &$200$&$2000$&$1024$\\
 				\# actor MLP units&$(128,128)$&$(32,32)$&$(64,64)$\\
 				\# critic MLP units&$(128,128)$&$(32,32)$&$(64,64)$\\
 				actor leanring rate &$3\times10^{-4}$&$3\times10^{-3}$&$10^{-2}$\\
 				critic leanring rate &$1\times10^{-3}$&$3\times10^{-3}$&$10^{-2}$\\
 				MLP activation &\multicolumn{3}{c}{ReLU}\\
 				optimizer&\multicolumn{3}{c}{Adam}\\
 				replay buffer size &-&-& $10^{6}$ \\
 				\midrule
 				\(T\)  &$200$&$100$&$400$\\
 				$p$  &\multicolumn{3}{c}{$0.8$} \\
 				VAE encoding space dimension &$1$&$1,5$&$1$\\
 				\# VAE encoder MLP units&\multicolumn{3}{c}{$(32,32)$}\\
 				\# VAE decoder MLP units&\multicolumn{3}{c}{$(32,32)$}\\
 				VAE learning rate&\multicolumn{3}{c}{$1\times10^{-4}$}\\
 				VAE training epochs&\multicolumn{3}{c}{$3$}\\
 				KDE bandwidth&\multicolumn{3}{c}{$0.05$}\\
 				KDE kernel&\multicolumn{3}{c}{Gaussian}\\
 				
 				\bottomrule
 			\end{tabular}
 		\end{footnotesize}
 	\end{table*}

\section{Hyperparameters}
 	
GENE and all the baselines work on a base RL algorithm. In each task, the hyperparameters of the base RL algorithm, such as batch size, learning rate, and discount factor, are all the same for fair comparison, which are summarized in Table~\ref{tab:hyperparameter}. The hyperparameters of GENE used in each task are also summarized in Table~\ref{tab:hyperparameter}. 
 	
\section{Maze}
In the 2D maze, the agent learns to navigate from a initial position to the target position within 200 timesteps. The initial position and the target position are indicated in Figure~\ref{fig:maze}. Only when the agent reaches the target, the agent receives a reward $+1$, otherwise always $0$. RND uses a target network with hidden layers $(32,32)$ to produce the representation with 8-dimension. The predictor network is trained to mimic the target network with the learning rate $10^{-4}$. The prediction error is used as the intrinsic reward to guide the agent to explore the novel state. As for RCG, we use the default setting $R_\text{min} = 0.1$ and $R_\text{max}=0.9$.
 	
\section{Maze Ant}
The ant \cite{duan2016benchmarking} within a square of  \({[0, 1]}^2\) and positioned at $P_a =  (0.125,0.875)$ tries to reach a goal with location $P_t =  (0.875,0.875)$ within $200$ timesteps. Only when the ant reaches the goal (\(||P_a-P_t||<0.125\)), it receives a reward of $+1$, otherwise always $0$. As for RCG, we use the default setting $R_\text{min} = 0.1$ and $R_\text{max}=0.9$. As for Goal GAN, $R_\text{min} = 0.1$ and $R_\text{max}=0.9$. The goal generator is an MLP with hidden layers $(128,128)$, and the goal discriminator is an MLP with hidden layers $(256,256)$. The generator takes as input 4-dimension noise sampled from the unit Gaussian distribution. The learning rate is $10^{-4}$.

\section{Cooperative Navigation}
 	
In Cooperative Navigation, there are the same number of landmarks (at four corners) and agents (at the center) in a square of \({[0, 1]}^2\). Only when each landmark is occupied by an agent (\(||P_a-P_t||<0.1\)), each agent receives a reward $+1$, otherwise always $0$. In MADDPG, the networks are updated every $50$ timesteps. As for RCG, we use the default setting $R_\text{min} = 0.1$ and $R_\text{max}=0.9$. As for HER, we use the final state of each episode as the additional goal for replay. 
 	
 	
 \end{document}